\documentclass[]{article}
\usepackage{proceed2e}

\usepackage{natbib}

\usepackage{times}

\usepackage{times}
\usepackage{graphicx} 
\usepackage{subfigure} 

\usepackage{natbib}

\usepackage{mydefs}

\usepackage{hyperref}

\usepackage{amsfonts,mathrsfs,amsmath}
\usepackage{times}
\usepackage{graphicx} 

\usepackage{wasysym,wrapfig,subfigure,epsfig,shapepar,array,colortbl,bm,url}
\usepackage[T1]{fontenc}

\definecolor{DarkGreen}   {rgb}{0,0.5,0}
\definecolor{DarkBlue}    {rgb}{0,0.0,0.5}
\definecolor{LightGray}   {rgb}{0.8,0.8,0.8}

\newcommand{\beq}{\vspace{0mm}\begin{equation}}
\newcommand{\eeq}{\vspace{0mm}\end{equation}}
\newcommand{\beqs}{\vspace{0mm}\begin{eqnarray}}
\newcommand{\eeqs}{\vspace{0mm}\end{eqnarray}}
\newcommand{\barr}{\begin{array}}
      \newcommand{\earr}{\end{array}}
\newcommand{\Amat}[0]{{{\bf A}}}

\newcommand{\Umat}[0]{{{\bf U}}}

\newcommand{\Xmat}[0]{{{\bf X}}}

\newcommand{\iv}{\boldsymbol{i}}

\newcommand{\uv}{\boldsymbol{u}}

\newcommand{\Ycal}{\mathcal{Y}}

\newcommand{\Bcal}{\mathcal{B}}
\newcommand{\Ocal}{\mathcal{O}}

\title{Zero-Truncated Poisson Tensor Factorization for Massive Binary Tensors}

\author{  {\bf Changwei Hu}\thanks{\ \ Equal contribution} \\
ECE Department \\
Duke University\\
Durham, NC 27708 \\
\And
{\bf Piyush Rai}$^*$  \\
ECE Department \\
Duke University\\
Durham, NC 27708 \\
\And
{\bf Lawrence Carin}   \\
ECE Department \\
Duke University\\
Durham, NC 27708 \\
}
\vspace{-2em}
\begin{document}

\maketitle

\begin{abstract}
We present a scalable Bayesian model for low-rank factorization of massive tensors with binary observations. The proposed model has the following key properties: (1) in contrast to the models based on the logistic or probit likelihood, using a zero-truncated Poisson likelihood for binary data allows our model to scale up in the number of \emph{ones} in the tensor, which is especially appealing for massive but sparse binary tensors; (2) side-information in form of binary pairwise relationships (e.g., an adjacency network) between objects in any tensor mode can also be leveraged, which can be especially useful in ``cold-start'' settings; and (3) the model admits simple Bayesian inference via batch, as well as \emph{online} MCMC; the latter allows scaling up even for \emph{dense} binary data (i.e., when the number of ones in the tensor/network is also massive). In addition, non-negative factor matrices in our model provide easy interpretability, and the tensor rank can be inferred from the data. We evaluate our model on several large-scale real-world binary tensors, achieving excellent computational scalability, and also demonstrate its usefulness in leveraging side-information provided in form of mode-network(s).
\end{abstract}

\vspace{-1em}
\section{INTRODUCTION}
\vspace{-0.5em}

With the recent surge in multiway, multirelational, or ``tensor'' data sets~\citep{nickel2011three,kang2012gigatensor}, learning algorithms that can extract useful knowledge from such data are becoming increasingly important. Tensor decomposition methods~\citep{kolda2009tensor} offer an attractive way to accomplish this task. Among tensor data, of particular interest are real-world \emph{binary} tensors, which are now ubiquitous in problems involving social networks, recommender systems, and knowledge bases, etc. For instance, in a knowledge base, predicate relations defined over the tuples (subjects, objects,verbs) can be represented in form of a binary three-way tensor~\citep{kang2012gigatensor}.

Usually, real-world binary tensors are massive (each dimension can be very large) but extremely sparse (very few ones in the tensor). For example, in a recommender system, each positive example (e.g., an item selected a set) implicitly creates several negative examples (items \emph{not} chosen). Likewise, in a knowledge base, the validity of one relation automatically implies invalidity of several other relations. In all these settings, the number of negative examples greatly overwhelms the number of positive examples. 

Unfortunately, binary tensor factorization methods~\citep{nickel2011three, xu2013bayesian,rai14tensor}, based on probit or logistic likelihood, scale poorly for massive binary tensors because these require evaluating the likelihood/loss-function on \emph{both} ones as well as zeros in the tensor. One possibility is to use heuristics such as \emph{undersampling} the zeros, but such heuristics usually result in less accurate solutions. Another alternative is to use the \emph{squared loss}~\citep{hidasi2012fast,nickel2012factorizing} as the model-fit criterion, which facilitates linear scalability in the number of ones in the tensor. However, such an approach can often lead to suboptimal results~\citep{ermisiterative} in practice.

It is therefore desirable to have methods that can perform efficient tensor decomposition for such data, ideally with a computational-complexity that depends only on the number of nonzeros (i.e., the ones) in the tensor, rather than the ``volume'' of the tensor. Motivated by this problem, we present a scalable Bayesian model for the Canonical PARAFAC (CP) tensor decomposition~\citep{kolda2009tensor}, with an inference-complexity that scales linearly in the number of ones in the tensor. Our model uses a zero-truncated Poisson likelihood for each binary observation in the tensor; this obviates the evaluation of the likelihoods for the zero entries. At the same time, the significant speed-up is not at the cost of sacrificing on the quality of the solution. As our experimental results show, the proposed likelihood model yields comparable or better results to logistic likelihood based models, while being an order of magnitude faster in its running-time on real-world binary tensors. Note that replacing the zero-truncated Poisson by the standard Poisson makes our model also readily applicable for count-valued tensors~\citep{chi2012tensors}; although, in this exposition, we will focus exclusively on binary tensors.

Often, side-information~\citep{acar2011all,beutel2014flexifact}, e.g., pairwise relationships (partially/fully observed), may also be available for objects in some of the tensor dimensions. For example, in addition to a binary tensor representing $\textsc{authors} \times \textsc{words} \times \textsc{venues}$ associations, the $\textsc{author} \times \textsc{author}$ co-authorship network may be available (at least for some pairs of authors). Such a network may be especially useful in ``cold-start'' settings where there is no data for some of the entities of a mode in the tensor (e.g., for some authors, there is no data in the tensor), but a network between entities in that mode may be available (See Fig~\ref{fig:coldstart} for an illustration). Our model allows leveraging such network(s), without a significant computational overhead, using the zero-truncated Poisson likelihood \emph{also} to model these binary pairwise relationships.

To facilitate efficient fully Bayesian inference, we develop easy-to-implement batch as well as \emph{online} MCMC inference; the latter is especially appealing for handling \emph{dense} binary data, i.e., when the number of ones in the tensor and/or the network is also massive. Another appealing aspect about the model is its interpretability; a Dirichlet prior on the columns of each factor matrix naturally imposes non-negativity. In addition, the rank of decomposition can be inferred from the data.

\vspace{-1em}
\section{CANONICAL PARAFAC (CP) TENSOR DECOMPOSITION}
\label{sec:cpdec}
\vspace{-0.5em}
The Canonical PARAFAC (CP) decomposition~\citep{kolda2009tensor} offers a way to express a tensor as a sum of rank-1 tensors. Each rank-1 tensor corresponds to a specific ``factor'' in the data. More specifically, the goal in CP decomposition is to decompose a tensor $\Ycal$ of size $n_1\times n_2 \times \cdots \times n_K$, with $n_{k}$ denoting the size of $\mathcal{Y}$ along the $k^{th}$ mode (or ``way'') of the tensor, into a set of $K$ factor matrices $\Umat^{(1)},\ldots,\Umat^{(K)}$ where $\Umat^{(k)} = [\uv_1^{(k)},\ldots,\uv_R^{(k)}], \ k = \{1,\ldots,K\}$, denotes the $n_k \times R$ factor matrix associated with mode $k$. 

In its most general form, CP decomposition expresses the tensor $\Ycal$ via a weighted sum of $R$ rank-1 tensors as 
      \vspace{-0.5em}
      \begin{equation}
      \Ycal \sim f(\sum_{r=1}^R \lambda_r . \uv_r^{(1)} \odot \ldots \odot \uv_r^{(K)})
      \label{eq:cp}
                  \vspace{-0.25em}
      \end{equation}
In the above, the form of the link-function $f$ depends on the type of data being modeled (e.g., $f$ can be Gaussian for real-valued, Bernoulli-logistic for binary-valued, Poisson for count-valued tensors). Here $\lambda_r$ is the weight associated with the $r^{th}$ rank-1 component, the $n_k\times 1$ column vector $\uv_r^{(k)}$ represents the $r^{th}$ latent factor of mode $k$, and $\odot$ denotes vector outer product. 

We use subscript $\iv = \{i_1,\ldots,i_K\}$ to denote the $K$-dimensional index of the $\iv$-th entry in the tensor $\Ycal$. Using this notation, the $\iv$-th entry of the tensor $\Ycal$ can be written as $y_{\vec{i}} \sim f(\sum_{r=1}^R \lambda_r \prod_{k=1}^K u_{i_k r}^{(k)})$.

\section{TRUNCATED POISSON TENSOR DECOMPOSITION FOR BINARY DATA}
\label{sec:bnbcp}
\vspace{-0.5em}
Our focus in this paper is on developing a probabilistic, fully Bayesian method for scalable low-rank decomposition of massive \emph{binary} tensors. As opposed to tensor decomposition models based on the logistic likelihood for binary data~\citep{xu2013bayesian,rai14tensor}, which require evaluation of the likelihood for both ones as well as zeros in the tensor, and thus can be computationally infeasible to run on massive binary tensors, our proposed model only requires the likelihood evaluations on the \emph{nonzero} (i.e., the ones) entries in the tensor, and can therefore easily scale to massive binary tensors. Our model is applicable to tensors of any order $K \geq 2$ (the case $K=2$ being a binary matrix).

Our model is based on a decomposition of the form given in Eq.~\ref{eq:cp}; however, instead of using a Bernoulli-logistic link $f$ to generate each binary observation $y_{\iv}$ in $\Ycal$, we assume an additional layer (Eq.~\ref{eq:addllayer}) which takes a \emph{latent} count-valued $y_{\iv}$ in $\Ycal$ and thresholds this latent count at one to generate the actual \emph{binary}-valued entry $b_{\iv}$ in the observed binary tensor, which we will denote by $\Bcal$: 
\beqs
b_{\iv}&=&{\boldsymbol 1}(y_{\iv}\ge 1) \label{eq:addllayer} \\ 
\Ycal &\sim& \text{Pois}(\sum_{r=1}^R \lambda_r . \uv_r^{(1)} \odot \ldots \odot \uv_r^{(K)})\label{eq:poistensor} \\
\uv^{(k)}_r &\sim& \text{Dir}(a^{(k)},\dots,a^{(k)})\label{eq:uprior}\\
\lambda_r&\sim& \text{Gamma}(g_r,\frac{p_r}{1-p_r})\label{eq:lambdaprior}\\
p_r&\sim& \text{Beta}(c\epsilon,c(1-\epsilon))\label{eq:prprior}
\vspace{-0.5em}
\eeqs
Marginalizing out $y_{\iv}$ from Eq.~\ref{eq:addllayer} leads to the following (equivalent) likelihood model
\vspace{-1em}
\begin{equation}
 b_{\iv} \sim \text{Bernoulli}(1-\exp(-\sum_{r=1}^R \lambda_r \prod_{k=1}^K u_{i_k r}^{(k)}))
 \label{eq:marglik}
\vspace{-0.5em} 
\end{equation}
Note that the thresholding in~(\ref{eq:addllayer}) looks similar to a probit model for binary data (which however thresholds a \emph{normal} at \emph{zero}); however, the probit model (just like the logistic model) also needs to evaluate the likelihood at the zeros, and can therefore be slow on massive binary data with lots of zeros. Likelihood models of the form (Eq.~\ref{eq:marglik}) have previously also been considered in work on statistical models of undirected networks~\citep{morup2011infinite,zhou2015infinite}. Interestingly, the form of the likelihood in~\ref{eq:marglik} also resembles the complementary log-log function~\cite{collett2002modelling,piegorsch1992complementary}, which is known to be a better model for imbalanced binary data than the logistic or probit likelihood, making it ideal for handling sparse binary tensors.  

The conditional posterior of the latent count $y_{\iv}$ is given by
\beq
\label{eq:yipost}
y_{\iv}|b_{\iv},\boldsymbol{\lambda},\{u_{i_k r}^{(k)}\}_{k=1}^K \sim b_{\iv}\cdot \text{Pois}_+(\sum_{r=1}^R \lambda_r \prod_{k=1}^K u_{i_k r}^{(k)})
\eeq
where $\text{Pois}_+(\cdot)$ is zero truncated Poisson distribution. Eq. (\ref{eq:yipost}) suggests that if $b_{\iv}=0$, then $y_{\iv}=0$ almost surely with probability one, which can lead to significant computational savings, if the tensor has a large number of zeros.  In addition, our model also enables leveraging a reparameterization (Section~\ref{sec:reparam}) of the Poisson distribution in terms of a multinomial, which allows us to obtain very simple Gibbs-sampling updates for the model parameters.

Note that the Dirichlet prior on the latent factors $\uv^{(k)}_r$ naturally imposes non-negativity constraints~\citep{chi2012tensors} on the factor matrices $\Umat^{(1)},\ldots,\Umat^{(K)}$. Moreover, since the columns $\uv^{(k)}_r$ of these factor matrices sums to 1, each $\uv^{(k)}_r$ can also be interpreted as a \emph{distribution} (e.g., a ``topic'') over the $n_k$ entities in mode $k$. Furthermore, the gamma-beta hierarchical construction~\citep{zhou2012NBPFA} of $\lambda_r$ (Eq~\ref{eq:lambdaprior} and ~\ref{eq:prprior}) allows inferring the rank of the tensor by setting an upper bound $R$ on the number of factors and inferring the appropriate number of factors by shrinking the coefficients $\lambda_r$'s to close to zero for the irrelevant factors. These aspects make our model interpretable as well as provide it the ability to do model selection (i.e., inferring the rank), in addition to being computationally efficient by focusing the computations only on the nonzero entries in the tensor $\Bcal$.

\subsection{LEVERAGING MODE NETWORKS}

\begin{figure}[!htbp]
\begin{center}
  \includegraphics[scale=0.2]{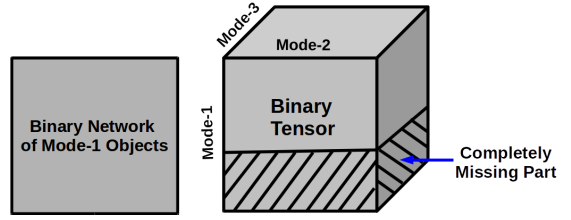}
  \caption{\small{Binary tensor with an associated binary network between objects in mode-1 of the tensor (in general, network for other modes may also be available). In the ``cold-start''setting as shown above, data along some of the tensor dimensions will be completely missing}}
	\label{fig:coldstart}
  \end{center}
	\vspace{-2em}
\end{figure}

Often, in addition to the binary tensor $\Bcal$, \emph{pairwise} relationships between entities in one or more tensor modes may be available in form of a symmetric binary network or an undirected graph. Leveraging such forms of side-information can be beneficial for tensor decomposition, especially if the amount of missing data in the main tensor $\Bcal$ is very high~\citep{acar2011all,beutel2014flexifact,rai2015leveraging}, and, even more importantly, in ``cold-start'' settings, where there is no data in the tensor for entities along some of the tensor mode(s), as shown in Fig~\ref{fig:coldstart}. In the absence of any side-information, the posterior distribution of the latent factors $\uv_{r}^{(k)}$ of such entities in that tensor mode would be the same as the prior (i.e., just a random draw). Leveraging the side-information (e.g., a network) helps avoid this. 

For entities of the $k$-th mode of tensor $\Bcal$, we assume a symmetric binary network $\Amat^{(k)} \in\{0,1\}^{n_k\times n_k}$, where $A_{i_kj_k}^{(k)}$ denotes the relationship between mode-$k$ entities $i_k$ and $j_k$. 

Just like our tensor decomposition model, we model the mode-$k$ network $\Amat^{(k)}$ as a weighted sum of rank-1 symmetric matrices, with a similar likelihood model as we use for the tensor observations. In particular, we assume a latent count $X_{i_kj_k}^{(k)}$ for each binary entry $A_{i_kj_k}^{(k)}$, and threshold it at one to generate $A_{i_kj_k}^{(k)}$
\beqs
A_{i_kj_k}^{(k)}&=&{\boldsymbol 1}(X_{i_kj_k}^{(k)}\ge 1)\\ 
\Xmat^{(k)} &\sim& \text{Pois}(\sum_{r=1}^R \beta_r . \uv_r^{(k)}\odot \uv_r^{(k)}) \\
\beta_r&\sim& \text{Gamma}(f_r,\frac{h_r}{1-h_r})\\
h_r&\sim& \text{Beta}(d\alpha,d(1-\alpha))
\eeqs
\vspace{-0.25em}
Note that since $\Amat^{(k)}$ is symmetric, only the upper (or lower) triangular portion needs to be considered, and moreover, just like in the case of the tensor $\Bcal$, due to the truncated Poisson construction, the likelihood at only the nonzero entries needs to be evaluated for this part as well. 

\subsection{REPARAMETERIZED POISSON DRAWS}
\label{sec:reparam}
To simplify posterior inference (Section~\ref{sec:infer}), we make use of two re-parameterizations of a Poisson draw~\citep{zhou2012NBPFA}. The first parameterization is to express each latent count variable $y_{\vec{i}}$ and $X_{i_kj_k}^{(k)}$ as a sum of another set of $R$ latent counts $\{\tilde{y}_{\vec{i}r}\}_{r=1}^R$ and $\{\tilde{X}_{i_kj_kr}^{(k)}\}_{r=1}^R$, respectively
\beqs
\label{eq:latentcounts}
y_{\vec{i}} = \sum_{r=1}^R \tilde{y}_{\vec{i}r}, \quad \tilde{y}_{\vec{i}r}\sim \text{Pois}(\lambda_r \prod_{k=1}^K u_{i_k r}^{(k)})\\
X_{i_kj_k}^{(k)} = \sum_{r=1}^R \tilde{X}_{i_kj_kr}^{(k)}, \quad \tilde{X}_{i_kj_kr}^{(k)}\sim \text{Pois}(\beta_r u_{i_k r}^{(k)}u_{j_k r}^{(k)})
\eeqs  
The second parameterization assumes that the latent counts $\{\tilde{y}_{\iv r}\}$ and $\tilde{X}_{i_kj_kr}^{(k)}$ are drawn from a multinomial
\beqs
&& \tilde{y}_{\vec{i}1},\ldots,\tilde{y}_{\vec{i}R} \sim \text{Mult}(y_{\vec{i}};\zeta_{\vec{i}1},\ldots,\zeta_{\vec{i}R}) \nonumber \\
&& \zeta_{\vec{i}r} = \frac{\lambda_r \prod_{k=1}^K u_{i_k r}^{(k)}}{\sum_{r=1}^R \lambda_r \prod_{k=1}^K u_{i_k r}^{(k)}}\label{eq:yirmultsample}\\
&& \tilde{X}_{i_kj_k1}^{(k)},\ldots,\tilde{X}_{i_kj_kR}^{(k)} \sim \text{Mult}(X_{i_kj_k}^{(k)};\kappa_{i_k j_k 1}^{(k)},\ldots,\kappa_{i_k j_k R}^{(k)}) \nonumber \\
&& \kappa_{i_k j_k r}^{(k)} = \frac{\beta_r u_{i_k r}^{(k)} u_{j_k r}^{(k)}}{\sum_{r=1}^R \beta_r u_{i_k r}^{(k)} u_{j_k r}^{(k)}}
\label{eq:xirmultsample}
\eeqs
As we show in Section~\ref{sec:infer}, these parameterizations enable us to exploit the gamma-Poisson as well as the Dirichlet-multinomial conjugacy to derive simple, closed-form Gibbs sampling updates for the model parameters.
      
\vspace{-0.5em}			
\section{INFERENCE}
\label{sec:infer}

Exact inference in the model is intractable and we resort to Markov Chain Monte Carlo (MCMC)~\citep{andrieu2003introduction} inference. In particular, the reparameterization discussed in Section~\ref{sec:reparam} allows us to derive simple Gibbs sampling updates for all the latent variables, except for the latent counts $y_{\iv}$, which are drawn from a truncated Poisson distribution via rejection sampling. As discussed earlier, the computational-cost for our inference method scales linearly w.r.t. the number of ones in the tensor (plus the number of nonzeros in the network, if side-information is used). This makes our method an order of magnitude faster than models based on logistic or probit likelihood for binary data~\citep{rai14tensor,xu2013bayesian}, without sacrificing on the quality of the results. The relative speed-up depends on the ratio of total volume of the tensor to the number of ones, which is given by $(\prod_{k=1}^K n_k)/\text{nnz}(\Bcal)$; here $\text{nnz}(\Bcal)$ denotes the number of nonzeros in the tensor.

In this section, we present both batch MCMC (Section~\ref{sec:batchmcmc}) as well as an online MCMC (Section~\ref{sec:onlinemcmc}) method for inference in our model. The online MCMC algorithm is based on the idea of Bayesian Conditional Density Filtering (CDF)~\citep{guhaniyogi2014bayesian}, and can lead to further speed-ups over the batch MCMC if the number of nonzeros in the tensor is also massive. The CDF algorithm provides an efficient way to perform online MCMC sampling using surrogate conditional sufficient statistics~\citep{guhaniyogi2014bayesian}.

For both batch MCMC and CDF based online MCMC, we provide the update equations, with and without the side-information, i.e., the mode network(s). For what follows, we define four quantities: $s_{j,r}^{(k)}=\sum_{\iv:i_k=j}\tilde{y}_{\iv r}$, $s_r=\sum_{\iv}\tilde{y}_{\iv,r}$, $v_{i_k,r}=\sum_{j_k}^{n_k}\tilde{X}_{i_kj_kr}^{(k)}$ and $v_r=\sum_{i_k}^{n_k}\sum_{j_k}^{n_k}\tilde{X}_{i_kj_kr}^{(k)}$, which denote aggregates computed using the latent counts $\tilde{y}_{\iv r}$ and $\tilde{X}_{i_kj_kr}^{(k)}$. These quantities will be used at various places in the description of the inference algorithms that we present here.

\subsection{BATCH MCMC INFERENCE}
\label{sec:batchmcmc}
\subsubsection{Tensor without Mode Network(s)}
\label{sec:batchmcmcwonetwork}
\textbf{Sampling $y_{\iv}$:} For each observation $b_{\iv}$ in the tensor, the latent count $y_{\iv}$ is sampled as
\beq
\label{eq:yisam}
y_{\iv}\sim b_{\iv}\cdot \text{Pois}_+(\sum_{r=1}^R \lambda_r \prod_{k=1}^K u_{i_k r}^{(k)})
\eeq
where $\text{Pois}_+(\cdot)$ is zero truncated Poisson distribution. Eq. (\ref{eq:yisam}) suggests that if $b_{\iv}=0$, then $y_{\iv}=0$ almost surely; and if $b_{\iv}=1$, then $y_{\iv}\sim \text{Pois}_+(\sum_{r=1}^R \lambda_r \prod_{k=1}^K u_{i_k r}^{(k)})$. Therefore the $y_{\iv}$'s only need to be sampled for the nonzero $b_{\iv}$'s.\\ 
\textbf{Sampling $\tilde{y}_{\iv r}$:} The latent counts $\{\tilde{y}_{\iv r}\}$ are sampled from a multinomial as Eq. (\ref{eq:yirmultsample}). Note that this also needs to be done only for the nonzero $b_{\iv}$'s.\\	
\textbf{Sampling $\uv_r^{(k)}$:} The columns of each factor matrix have a Dirichlet posterior, and are sampled as
\beq
\uv_r^{(k)}\sim \text{Dir}(a^{(k)}+ s_{1,r}^{(k)},a^{(k)}+s_{2,r}^{(k)},\ldots,a^{(k)}+s_{n_k,r}^{(k)})
\label{eq:factorswonetwork}
\eeq		
\textbf{Sampling $p_r$:}
Using the fact that $s_r=\sum_{\iv}\tilde{y}_{\iv,r}$ and marginalizing over the $u_{i_k r}^{(k)}$'s in (\ref{eq:latentcounts}), we have $s_r \sim \text{Pois}(\lambda_r)$. Using this, along with (\ref{eq:lambdaprior}), we can express $s_r$ using a negative binomial distribution, i.e., $s_r\sim \text{NB}(g_r,p_r)$. Due to the conjugacy between negative binomial and beta, we can then sample $p_r$ as 
\beq
p_r\sim \text{Beta}(c\epsilon+s_r,c(1-\epsilon)+g_r)
\eeq 
\textbf{Sampling  $\lambda_r$:} Again using the fact that $s_r \sim \text{Pois}(\lambda_r)$ and (\ref{eq:lambdaprior}), we have 
\vspace{-0.5em}
\beq
\lambda_r\sim \text{Gamma}(g_r+s_r,p_r)
\eeq
As can be observed, when updating $\uv_r^{(k)}$, $p_r$ and $\lambda_r$, the latent counts $y_{\iv}$'s and $\tilde{y}_{\iv r}$ corresponding to zero entries in $\Bcal$ are all equal to zero, and have no contribution to sufficient statistics $s_{j,r}^{(k)}$ and $s_r$. Therefore, only the nonzero entries in tensor need to be considered in the computations. 

\subsubsection{Tensor with Mode Network(s)}
\label{sec:batchmcmcwnetwork}
In the presence of mode network(s), the update equations for the latent variables $p_r$, $\lambda_r$, $\tilde{y}_{\iv r}$ and $y_{\iv}$, that are associated solely with the binary tensor $\Bcal$, remain unchanged, and can be sampled as described in Section~\ref{sec:batchmcmcwonetwork}. We however need to sample the additional latent variables associated with mode-$k$ network $\mat{A}^{(k)}$, and the latent factors $\uv_r^{(k)}$ of mode-$k$ that are shared by the binary tensor $\Bcal$ as well as the mode-$k$ network.

\textbf{Sampling $X_{i_kj_k}^{(k)}$:} The latent counts $X_{i_kj_k}^{(k)}$ are sampled as
\beq
\label{eq:xijsam}
X_{i_kj_k}^{(k)}\sim A_{i_kj_k}^{(k)}\cdot \text{Pois}_+(\sum_{r=1}^R \beta_r u_{i_k r}^{(k)} u_{j_k r}^{(k)})
\eeq
This only needs to be done for the nonzero entries in $\mat{A}^{(k)}$.
	
\textbf{Sampling $\tilde{X}_{i_kj_kr}$:} The latent counts $\tilde{X}_{i_kj_kr}$ are sampled from a multinomial as equation (\ref{eq:xirmultsample}). This also only needs to be done for the nonzero entries in $\mat{A}^{(k)}$.

\textbf{Sampling $\uv_r^{(k)}$:} The columns of each factor matrix have a Dirichlet posterior, and are sampled as
\beq
\uv_r^{(k)}\sim \text{Dir}(a^{(k)}+ s_{1,r}^{(k)}+v_{1,r},\ldots,a^{(k)}+s_{n_k,r}^{(k)}+v_{n_k,r})
\label{eq:factorswnetwork}
\eeq	
Note that in the absence of the mode-$k$ network, the terms $v_{.,r}$ go away and Eq.~\ref{eq:factorswnetwork} simply reduces to Eq.~\ref{eq:factorswonetwork}.

\textbf{Sampling $h_r$:} $h_r\sim \text{Beta}(d\alpha+v_r,d(1-\alpha)+f_r)$.

\textbf{Sampling  $\beta_r$:} $\beta_r\sim \text{Gamma}(f_r+v_r,h_r)$.
\subsubsection{Per-iteration time-complexity} 
\label{sec:timecomplexity}
For the binary tensor $\Bcal$, computing each $\zeta_{\vec{i}r}$ (Eq.~\ref{eq:yirmultsample}) takes $\Ocal(K)$ time and therefore computing all the $\{\zeta_{\vec{i}r}\}$ takes $\Ocal(\text{nnz}(\Bcal)RK)$ time. Likewise, for the binary mode-$k$ network $\Amat^{(k)}$, computing all the $\{\kappa_{i_k j_k r}^{(k)}\}$ (Eq.~\ref{eq:xirmultsample}) takes $\Ocal(\text{nnz}(\Amat^{(k)})R)$ time. These are the most dominant computations in each iteration of our MCMC procedure; updating each $\uv_r^{(k)}$ takes $\Ocal(n_k)$ time and updating $\{p_r,h_r\}_{r=1}^R$ and $\{\lambda_r,\beta_r\}_{r=1}^R$ takes $\Ocal(R)$ time each. Therefore, the per-iteration time-complexity of our batch MCMC method is $\Ocal(\text{nnz}(\Bcal)RK + \text{nnz}(\Amat^{(k)})R)$. The linear dependence on $\text{nnz}(\Bcal), \text{nnz}(\Amat^{(k)}), R$ and $K$ suggests that even massive, sparse binary tensors and mode network(s) can be handled easily even by our simple batch MCMC implementation. Also note that our model scales linearly even w.r.t. $R$, unlike most other methods~\citep{ermisiterative,rai14tensor} that have \emph{quadratic} dependence on $R$. 

The above computations can be further accelerated using a distributed/multi-core setting; we leave this for future work. In Section~\ref{sec:onlinemcmc}, however, we present an \emph{online} MCMC method based on the idea of Bayesian Conditional Density Filtering~\citep{guhaniyogi2014bayesian}, which leads to further speed-ups, even in single-machine settings.
	
\subsection{ONLINE MCMC INFERENCE}
\label{sec:onlinemcmc}

We develop an efficient online MCMC sampler for the model, leveraging ideas from the Conditional Density Filtering (CDF)~\cite{guhaniyogi2014bayesian}. The CDF algorithm for our model selects a minibatch of the tensor (and mode network, if the side-information is available) entries at each iteration, samples the model parameters from the posterior, and updates the sufficient statistics $s_{j,r}^{(k)}$, $s_r$, $v_{i_k,r}$ and $v_r$ using the data from the current minibatch.

\subsubsection{Tensor without Mode Network(s)}

We first provide the update equations for the case when there is no side-information (mode network). Denote $I_t$ as indices of entries of tensor $\Bcal$ from the minibatch selected at iteration $t$. The CDF algorithm at iteration $t$ proceeds as:

\textbf{Sampling $y_{\iv}$:} For all $\iv\in I_t$, sample $y_{\iv}$ according to equation (\ref{eq:yisam}); like in the batch MCMC case, the sampling only needs to be done for the nonzero $b_{\iv}$'s.

\textbf{Sampling $\tilde{y}_{\iv r}$:} For all $\iv\in I_t$, sample the latent counts $\tilde{y}_{\iv r(\iv\in I_t)}$ using (\ref{eq:yirmultsample}), again only for the nonzero $b_{\iv}$'s.

\textbf{Updating the conditional sufficient statistics:} Update the conditional sufficient statistics $s_{j,r}^{(k)}$ as $s_{j,r}^{(k,t)} = s_{j,r}^{(k,t-1)} + \sum_{\iv \in I_t:i_k=j}\tilde{y}_{\iv r}$ and update $s_r$ as $s_r^{(t)} = s_r^{(t-1)} + \sum_{\iv\in I_t}\tilde{y}_{\iv,r}$. These updates basically add to the old sufficient statistics, the contributions from the data in the current minibatch. In practice, we also \emph{reweight} these sufficient statistics by the ratio of the total number of ones in $\Bcal$ and the minibatch size, so that they represent the average statistics over the entire tensor. This reweighting is akin to the way average gradients are computed in stochastic variational inference methods~\citep{hoffman2013stochastic}.

\textbf{Updating $\uv_r^{(k)}, p_r, \lambda_r$:} Using the following conditionals, draw $M$ samples $\{\uv_r^{(k,m)}, p_r^{(m)}, \lambda_r^{(m)}\}_{m=1}^M$
\begin{eqnarray}
\uv_r^{(k)} &\sim& \text{Dir}(a^{(k)}+s_{1,r}^{(k,t)},\ldots,a^{(k)}+s_{n_k,r}^{(k,t)}) \\
p_r &\sim& \text{Beta}(c\epsilon+s_r^{(t)},c(1-\epsilon)+g_r) \\
\lambda_r &\sim& \text{Gamma}(g_r+s_r^{(t)},p_r)
\end{eqnarray}
and either store the sample averages of $\uv_r^{(k)}, p_r$, and $\lambda_r$, or their analytic means to use for the next CDF iteration~\citep{guhaniyogi2014bayesian}.

\subsubsection{Tensor with Mode Network(s)}

For all the latent variables associated solely with the tensor $\Bcal$, the sampling equations for the CDF algorithm in the presence of mode network(s) remain unchanged as the previous case with no network.
In the presence of the mode network, the additional latent variables include the sufficient statistics $v_{i_k,r}$ and $v_r$, and these need to be updated in each CDF iteration. 

Denote $J_t$ as indices of entries selected from the mode-$k$ network $\mat{A}^{(k)}$ in iteration $t$. The update equations for the latent variables that depend on $\mat{A}^{(k)}$ are as follows:

\textbf{Sampling $X_{i_kj_k}$:} For $(i_k,j_k)\in J_t$, latent count $X_{i_kj_k}$ is sampled using Eq.~(\ref{eq:xijsam}).

\textbf{Sampling $\tilde{X}_{i_kj_kr}$:} For $(i_k,j_k)\in J_t$, latent counts $\tilde{X}_{i_kj_kr}$ are sampled from a multinomial using Eq.~(\ref{eq:xirmultsample}).

\textbf{Updating the conditional sufficient statistics:} Update the sufficient statistics associated with the mode-$k$ network as $v_{i_k,r}^{(t)} = v_{i_k,r}^{(t-1)} + \sum_{j_k,(i_k,j_k)\in J_t}^{n_k}\tilde{X}_{i_kj_kr}$ and $v_r^{(t)} = v_r^{(t-1)} + \sum_{i_k}^{n_k}\sum_{j_k,(i_k,j_k)\in J_t}^{n_k}\tilde{X}_{i_kj_kr}$. Just like the way we update the tensor sufficient statistics $s_{j,r}^{(k)}$ and $s_r$, we reweight these mode-$k$ sufficient statistics by the ratio of the total number of ones in $\mat{A}^{(k)}$ and the minibatch size, so that they represent the average statistics over the entire mode-$k$ network.
	
\textbf{Updating $\uv_r^{(k)}$, $h_r$, $\beta_r$:} Using the following conditionals, draw $M$ samples $\{\uv_r^{(k,m)}, h_r^{(m)}, \beta_r^{(m)}\}_{m=1}^M$. We draw $\uv_r^{(k)} \sim \text{Dir}(a^{(k)}+s_{1,r}^{(k,t)}+v_{1,r}^{(t)},\ldots,a^{(k)}+s_{n_k,r}^{(k,t)}+v_{n_k,r}^{(t)})$, and $h_r$ and $\beta_r$ as
\begin{eqnarray}
h_r &\sim& \text{Beta}(d\alpha+v_r^{(t)},d(1-\alpha)+f_r) \nonumber\\
\beta_r &\sim& \text{Gamma}(f_r+v_r^{(t)},h_r)
\end{eqnarray}
and either store the sample averages of $\uv_r^{(k)}$, $h_r$, $\beta_r$, or their analytic means to use for the next CDF iteration.\\
 \vspace{-1em}     
\subsubsection{Per-iteration time-complexity} 

The per-iteration time-complexity of the CDF based online MCMC is linear in the number of nonzeros in each minibatch (as opposed to the batch MCMC where it depends on the number of nonzeros in the \emph{entire} tensor and network). Therefore the online MCMC is attractive for \emph{dense} binary data, where the number of nonzeros in the tensor/network is also massive; using a big-enough minibatch size (that fits in the main memory and/or can be processed in each iteration in a reasonable amount of time), the online MCMC inference allows applying our model on such dense binary data as well, which may potentially have several billions of nonzero entries.

\vspace{-0.5em}
\section{RELATED WORK}
\label{sec:relwork}
\vspace{-0.5em}

With the increasing prevalence of structured databases, social networks, and (multi)relational data, tensor decomposition methods are becoming increasingly popular for extracting knowledge and doing predictive analytics on such data~\citep{bordes2011learning,nickel2012factorizing,kang2012gigatensor}. As the size of these data sets continues to grow, there has been a pressing need to design tensor factorization methods that can scale to massive tensor data.

For low-rank factorization of \emph{binary} tensors, methods based on logistic and probit likelihood for the binary data have been proposed~\citep{jenatton2012latent,london2013multi,rai14tensor,xu2013bayesian}. However, these methods are not suited for massive binary tensors where the number of observations (which mostly consist of zeros, if the tensor is also sparse) could easily be millions or even billions~\citep{inah2015haten2}. As a heuristic, these methods rely on subsampling~\citep{rai14tensor} or partitioning the tensor~\citep{zhescalable}, to select a manageable number entries before performing the tensor decomposition, or alternatively going for a distributed setting~\citep{zhe2013dintucker}.

In the context of tensor factorization, to the best of our knowledge, the only method (and one that is closest in spirit to our work) that scales linearly w.r.t. the number of ones in the tensor is~\citep{ermisiterative}. Their work explored quadratic loss (and its variations) as a surrogate to the logistic loss and proposed a method (\texttt{Quad-Approx}) with a per-iteration complexity $\Ocal(\text{nnz}(\Bcal)R + R^2\sum_{k=1}^K n_k)$. Note that its dependence on $R$ is quadratic as opposed to our method which is also linear in $R$. They also proposed variations based on piecewise quadratic approximations; however, as reported in their experiments~\citep{ermisiterative}, these variations were found to be about twice as slow than their basic \texttt{Quad-Approx} method~\citep{ermisiterative}. Moreover, their methods (and the various other methods discussed in this section) have several other key differences from our proposed model: (1) our model naturally imposes non-negativity on the factor matrices; (2) $R$ can be inferred from data; (3) our method provides a fully Bayesian treatment; (4) in contrast to their method, which operates in a batch setting, the online MCMC inference allows our model to scale to even bigger problems, where the number of nonzeros could also be massive; and (5) our model also allows incorporating (fully or partially observed) mode-networks as a rich source of side-information.

In another recent work~\citep{zhou2015infinite}, a similar zero-truncated Poisson construction, as ours, was proposed for \emph{edge-partioning} based network clustering, allowing the proposed model to scale in terms of the number of edges in the network. Our model, on the other hand, is more general and can be applied to multiway binary tensor data, with an optionally available binary network as a potential source of side-information. Moreover, the Dirichlet prior on the factor matrices, its reparametrizations (Section~\ref{sec:reparam}), and the online MCMC inference lead to a highly scalable framework for tensor decomposition with side-information.

Another line of work on scaling up tensor factorization methods involves developing distributed and parallel methods~\citep{kang2012gigatensor,inah2015haten2,papalexakis2012parcube,beutel2014flexifact}. Most of these methods, however, have one or more of the following limitations: (1) these methods lack a proper generative model of the data, which is simply assumed to be real-valued and the optimization objective is based on minimizing the Frobenius norm of the tensor reconstruction error, which may not be suitable for binary data; (2) these methods usually assume a parallel or distributed setting, and therefore are not feasible to run on a single machine; (3) missing data cannot be easily handled/predicted; and (4) the rank of the decomposition needs to be chosen via cross-validation.

Leveraging sources of side-information for tensor factorization has also been gaining a lot of attention recently. However, most of these methods cannot scale easily to massive tensors~\citep{acar2011all,rai2015leveraging}, or have to rely on parallel or distributed computing infrastructures~\citep{beutel2014flexifact}. In contrast, our model, by the virtue of its scalability that only depends on the number of nonzero entries in the tensor and/or the mode network, easily allows it to scale to massive binary tensors, with or without mode-network based side-information.

\section{EXPERIMENTS}
\label{sec:expt}
\vspace{-1em}
We report experimental results for our model on a wide range of real-world binary tensors (with and without mode-network based side-information), and compare it with several baselines for binary tensor factorization. We use the following data sets for our experiments:

\begin{table*}[!htbp]
\scriptsize
      \caption{\small{Tensor completion accuracies in terms of AUC-ROC scores. Results are averaged over 10 splits of training and test data. Note: (1) Bayesian CP was infeasible to run on the Scholars and Facebook data; (2) Due to the lack of publicly available code for \textbf{Quad-App} and \textbf{PQ-QuadApp}, we only report its results on Kinship, UMLS, and MovieLens data (results taken from~\citep{ermisiterative}).}}
      \label{tab:binary}  
\center
      \begin{tabular}{|c|c|c|c|c|c|c|}
     	\hline					
	  & \bf{Kinship}  & \bf{UMLS} & \bf{Movielens}  & \bf{DBLP} & \bf{Scholars}  & \bf{Facebook}\\  
    	\hline    
    	\bf{Quad-App~\citep{ermisiterative}} & 0.8193 & 0.8205 & 0.8511  & - & - & -\\ \hline
	\bf{PW-QuadApp~\citep{ermisiterative}} & 0.9213  & 0.9387 & 0.9490  & - & - & - \\ \hline 			
	\bf{Bayesian-Logistic-CP~\citep{rai14tensor}} & \bf{0.9865} & \bf{0.9965} & 0.9799 & 0.9307 & - & -\\ \hline	
	\bf{ZTP-CP (Batch MCMC)} & 0.9674 & 0.9938 & \bf{0.9895} & \bf{0.9759} & \bf{0.9959} & 0.9830 \\ \hline
	\bf{ZTP-CP (Online MCMC)} & 0.9628 & 0.9936 & 0.9841 & 0.9743& 0.9958 & \bf{0.9844} \\ \hline 
      \end{tabular}
      \vspace{-0.5em}   
\end{table*}

\begin{itemize}
 \item \textbf{Kinship:} This is a binary tensor of size $104 \times 104 \times 26$, representing 26 types of relationships between 104 members of a tribe~\citep{nickel2011three}. The tensor has about 3.8\% nonzeros.
 \item \textbf{UMLS:} This is a binary tensor of size $135 \times 135 \times 49$ representing 56 types of verb relations between 135 high-level concepts~\citep{nickel2011three}. The tensor has about 0.8\% nonzeros.
 \item \textbf{Movielens:} This is a binary \emph{matrix} (two-way tensor) of size $943 \times 1682$ representing the binary ratings (thumbs-up or thumbs-down) by 943 users on 1682 movies~\footnote{\url{http://grouplens.org/datasets/movielens/}}. This data set has a total of 100,000 ones.
 \item \textbf{DBLP:} This is a binary tensor of size $10,000 \times 200 \times 10,000$ representing (author-conference-keyword) relations~\citep{zhescalable}. This tensor has only about 0.001\% nonzeros, and is an ideal example of a massive but sparse binary tensor.
 \item \textbf{Scholars:} This is a binary tensor of size $2370  \times 8663 \times 4066$, constructed from a database of research paper abstracts published by researchers at Duke University; the three tensor modes correspond to authors, words, and publication venues, respectively. Just like the DBLP data, this tensor is also massive but extremely sparse with only about 0.002\% nonzeros. In addition, the co-authorship network (i.e., who has written papers with whom) is also available, which we use as a source of side-information, and use this network to experiment with the \emph{cold-start} setting (i.e., when the main tensor has no information about some authors).
 \item \textbf{Facebook:} The Facebook data is a binary tensor of size $63731 \times 63730 \times 1837$ with the three modes representing wall-owner, poster, and days~\citep{papalexakis2013scoup}. This tensor has only 737498 nonzeros. In addition to the binary tensor, the social network (friendship-links) between users is also given in form of a symmetric binary matrix of size $63731 \times 63731$, which has 1634180 nonzeros. We use the network to experiment with the cold-start setting.
\end{itemize}

We use all the 6 data sets for the tensor completion experiments (Section~\ref{sec:tendecomp}). We also use the Scholars and Faceboook data in the cold-start setting, where we experiment on the tensor completion task, leveraging the mode-network based side-information (Section~\ref{sec:tens-net}).

The set of experiments we perform includes: (1) binary tensor completion (Section~\ref{sec:tendecomp}) using only the tensor data; (2) scalability behavior of our model (both batch as well as online MCMC) in terms of tensor completion accuracy vs run-time (Section~\ref{sec:scalability}); we compare our model with Bayesian CP based on logistic-likelihood~\citep{rai14tensor}; (3) a qualitative analysis of our results using a \emph{multiway} topic modeling experiment (Section~\ref{sec:multiwaytopic}) on the Scholars data, with the entities being authors, words, and publication venues; and (4) leveraging the mode network for tensor completion in the cold-start setting (Section~\ref{sec:tens-net}); for this experiment, we also demonstrate how leveraging the network leads to improved qualitative results in the multiway topic modeling problem.

In the experiments, we refer to our model as \textbf{ZTP-CP} (Zero-Truncated Poisson based CP decomposition). We compare \textbf{ZTP-CP} (using both batch MCMC as well as online MCMC inference) with the following baselines: (1) the quadratic loss minimization (\textbf{Quad-App}) proposed in~\citep{ermisiterative}; (2) the refined piecewise quadratic approximation algorithm (\textbf{PW-QuadApp})~\citep{ermisiterative}; and (3) \textbf{Bayesian CP} decomposition based on logistic likelihood for binary data~\citep{rai14tensor}. 

\textbf{Experimental settings:} All experiments are done on a standard desktop
computer with Intel i7 3.4GHz processor and 24GB RAM. Unless specified otherwise, the MCMC inference was run for 1000 iterations with 500 burn-in iterations. The online MCMC algorithm was also run for the same number of iterations, with minibatch size equal to one-tenth of the number of nonzeros in the training data. For all the data sets, except Scholars and Facebook, we use $R=20$ (also note that our model has the ability to prune the unnecessary factors by shrinking the corresponding $\lambda_r$ to zero). For Scholars and Facebook data, we set $R=100$. The hyperparameters $g_r, f_r$ were set to 0.1, and $\epsilon$ and $\alpha$ are set to $1/R$, which worked well in our experiments.

\vspace{-1em}	
\subsection{TENSOR COMPLETION}
\label{sec:tendecomp}

In Table~\ref{tab:binary}, we report the results on the tensor completion task (in terms of the AUC-ROC - the area under the ROC curve). For this experiment, although available, we do not use the mode network for the Scholars and the Facebook data; only the binary tensor is used (the results when also using the network are reported in Section~\ref{sec:tens-net}). For each data set, we randomly select 90\% of the tensor observations as the training data and evaluate each model on the remaining 10\% observations used as the held-out data. 

Since the code for \textbf{Quad-App} and \textbf{PW-QuadApp} baselines (both proposed in~\citep{ermisiterative}) is not publicly available, we are only able to report the results for the Kinship, UMLS, and MovieLens data set (using the results reported in~\citep{ermisiterative}). For Bayesian CP~\citep{rai14tensor}, we use the code provided by the authors. Moreover, the Bayesian CP baseline was found infeasible to run on the Scholars and Facebook data (both of which are massive tensors), so we are unable to report those results. For fairness, on Kinship, UMLS, and MovieLens data, we use the same experimental settings for all the methods as used by~\citep{ermisiterative}.

As shown in Table~\ref{tab:binary}, our model outperforms \textbf{Quad-App} and \textbf{PW-QuadApp} in terms of the tensor-completion accuracies, and performs comparably or better than \textbf{Bayesian CP}, while being an order of magnitude faster (Section~\ref{sec:scalability} shows the results on running times).

\vspace{-1.25em}	
\subsection{SCALABILITY}
\label{sec:scalability}
\begin{figure}[!htbp]
		\begin{center}
			\centerline{
				\includegraphics[scale=0.23]{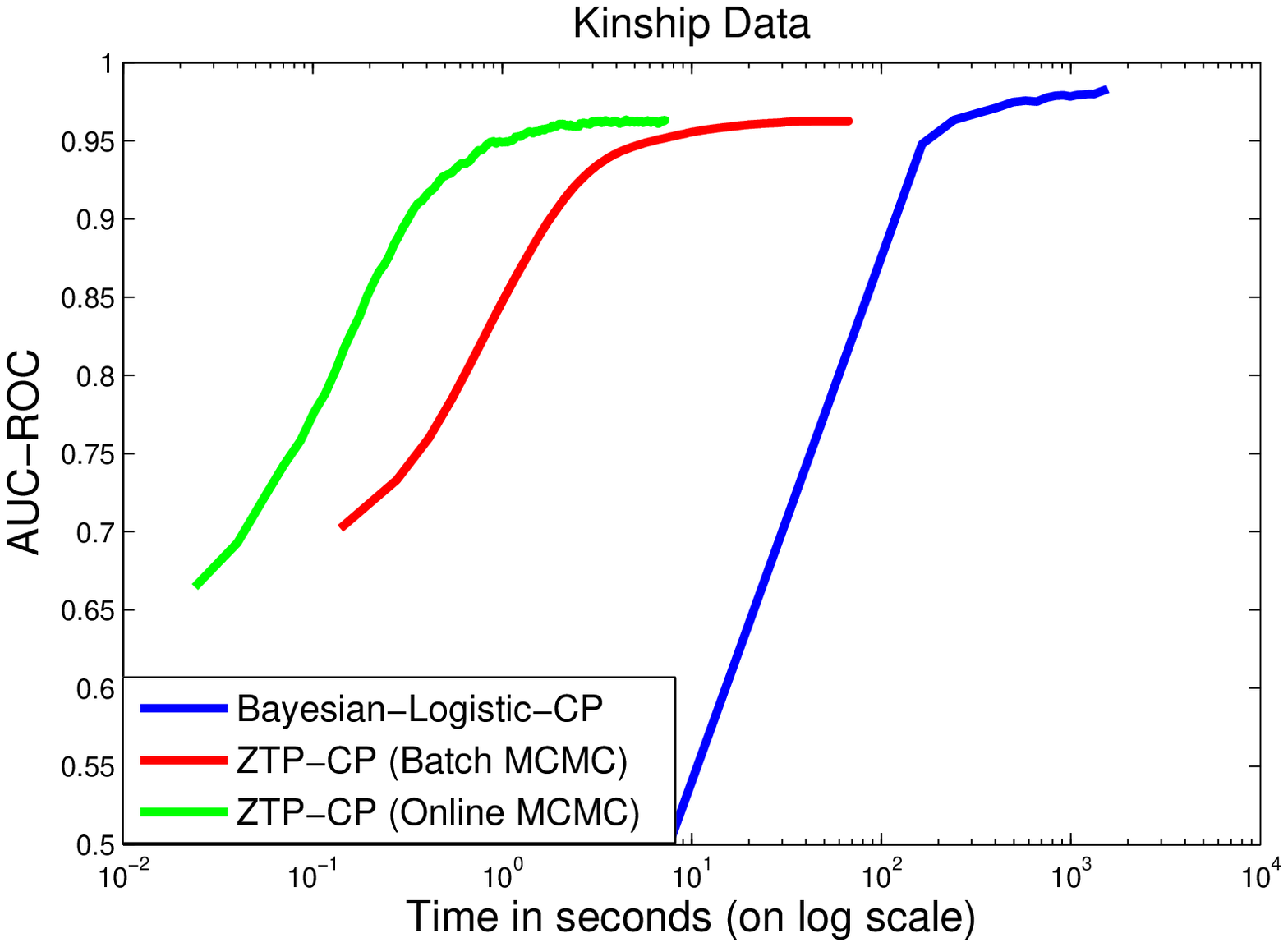}
				\includegraphics[scale=0.23]{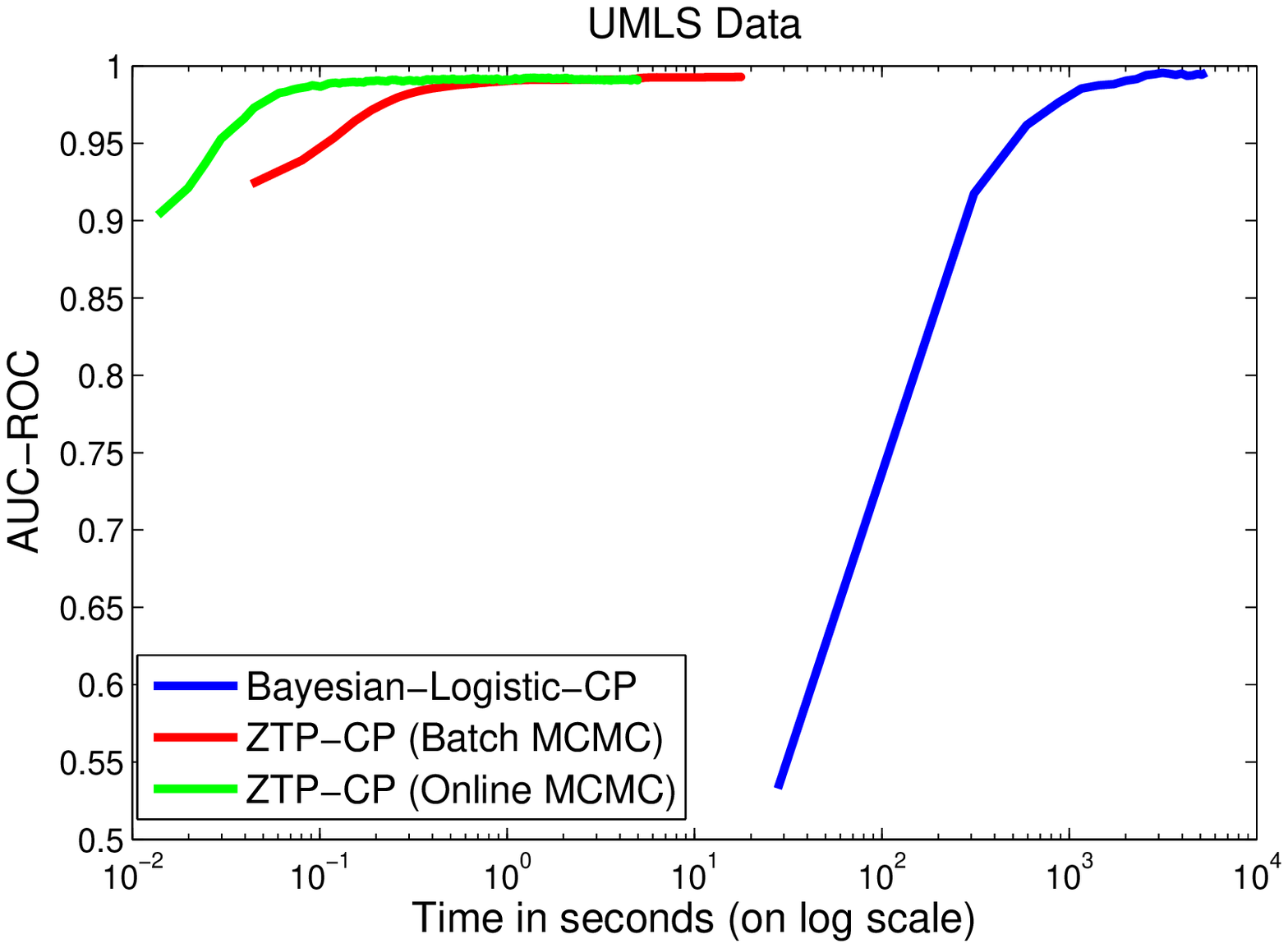}
			}
			\vspace{0.5em}
			\centerline{
				\includegraphics[scale=0.21]{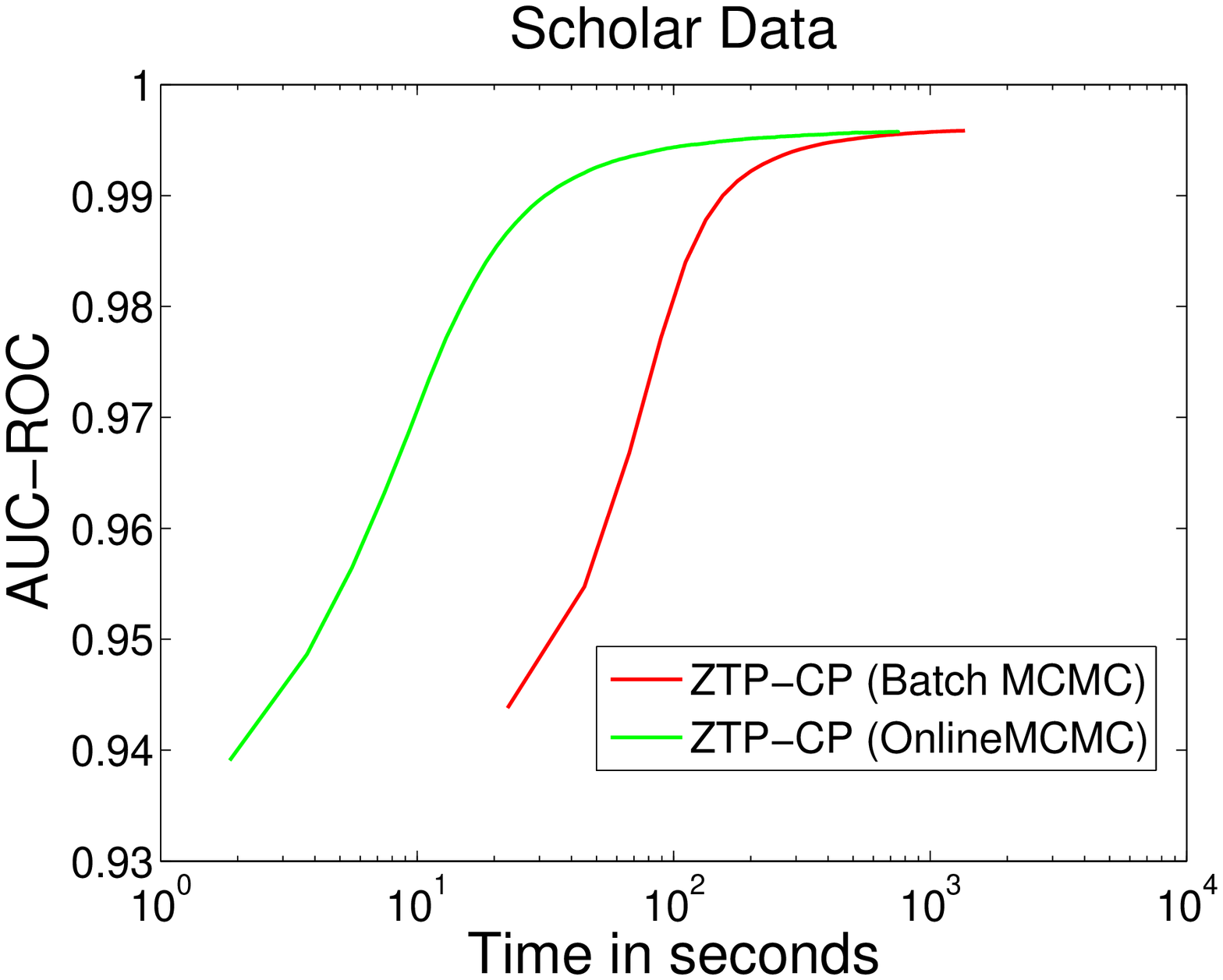}
				\includegraphics[scale=0.21]{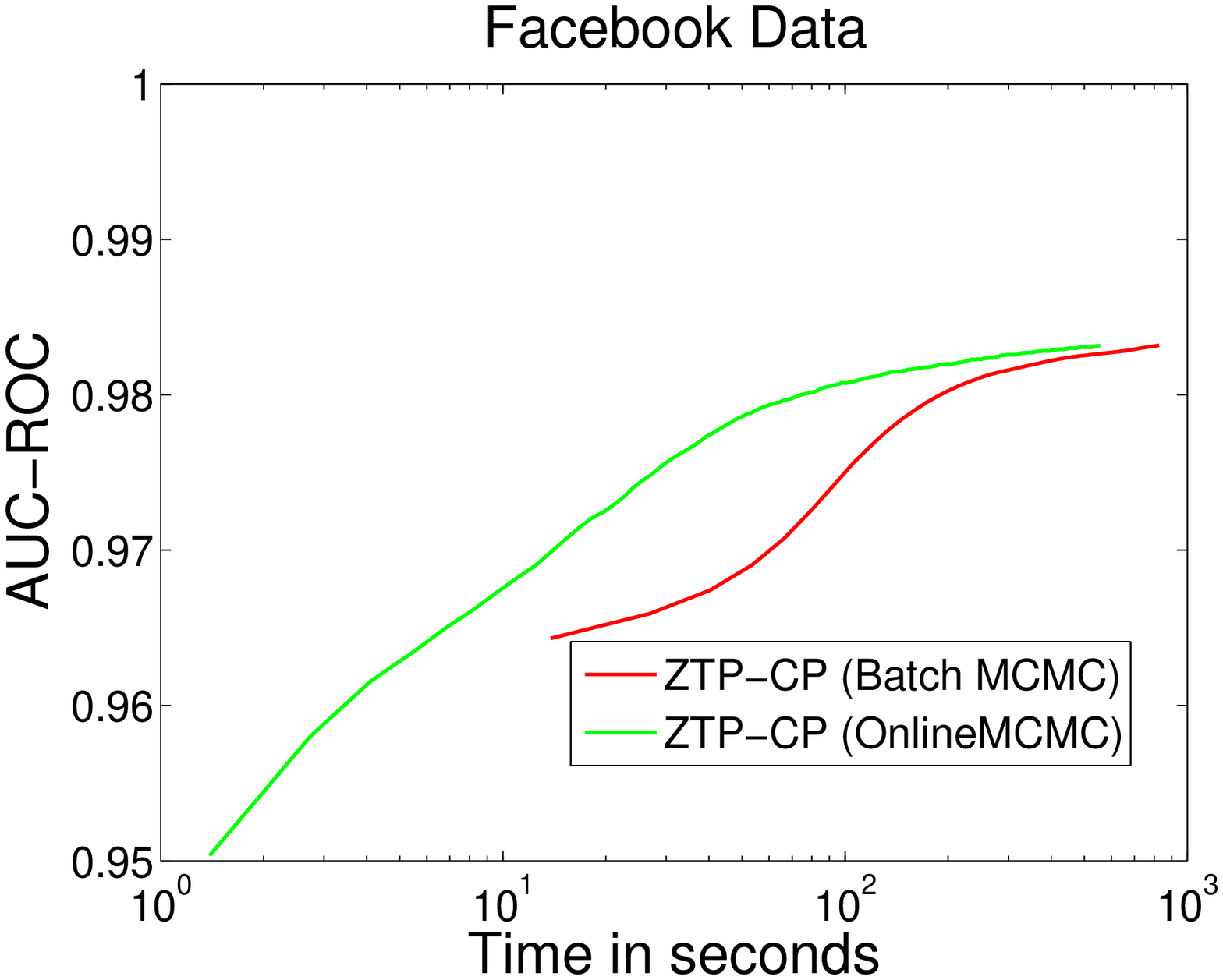}
			}
			\vskip -0.5em
			\caption{\small{Running time (log-scale) comparison of various methods on Kinship (top left), UMLS (top right), Scholars (bottom left), and Facebook (bottom right) datasets.}}
			\label{fig:timecomp}
		\end{center}
		\vskip -0.4in
	\end{figure}

\vspace{-0.75em}
We next compare our model with Bayesian CP~\citep{rai14tensor} in terms of the running times vs tensor completion accuracy on Kinship and UMLS data sets. As shown in Fig.~\ref{fig:timecomp} (top-row), our model (batch as well as online MCMC) runs/converges an order of magnitude faster than Bayesian CP in terms of running time. On Scholars and Facebook, since Bayesian CP was infeasible to run, we are only able to show the results (Fig.~\ref{fig:timecomp}, bottom-row) for our model, with batch MCMC and online MCMC inference. On all the data sets, the online MCMC runs/converges faster than the batch MCMC. 

We would like to note that, although the model proposed in~\citep{ermisiterative} also scales linearly~\footnote{Although~\citep{ermisiterative} reported run times on Kinship and UMLS data sets, those number are not directly comparable with our run times reported here (due to possibly different machine configuration, which they do not specify in the paper).} in the number of ones in the tensor, the per-iteration time-complexity of our model, which is linear in \emph{both} $\text{nnz}(\Bcal)$ as well as rank $R$, is better than the model proposed in~\citep{ermisiterative} (which has \emph{quadratic} dependence on $R$). Moreover, the tensor completion results of our model (shown in Table~\ref{tab:binary}) on these data sets are better than the ones reported in~\citep{ermisiterative}.

\begin{table*}[!htbp]
		\caption{\small{For the Scholars data, the most probable words in topics related to evolutionary biology (Evo Bio), medical imaging (Med Imag), machine learning/signal processing(ML/SP) and oncology, and top ranked venues in ML/SP}}
		\label{tab:topic}
		\vskip -0.15in
		\begin{center}
			\begin{scriptsize}
				\begin{sc}
					\begin{tabular}{llll|l}
						\hline
						Evo Bio & Med Imag & ML/SP & Oncology & Top Venues in ML/SP\\
						\hline
						species     & imaging     & bayesian     & radiation    & ICASSP\\
						selection   & contrast    & algorithm    & radiotherapy & JASA\\
						genetic     & computed    & sampling     & stage        & ICML\\
						evolution   & resonance   & features     & tumor        & IEEE trans img proc\\
						populations & dose        & process      & survival     & NIPS\\
						evolutionary& tomography  & sparse       & lung         & Compu stat data analy\\
						gene        & magnetic    & nonparametric& chemotherapy & Biometrics\\
						variation   & image       & gibbs        & treated      & Bayesian analysis\\
						plants      & quality     & parameters   & toxicity     & JMLR\\
						natural     & diagnostic  & inference    & oncology     & IEEE trans. inf. theory\\
						\hline
					\end{tabular}
				\end{sc}
			\end{scriptsize}
		\end{center}
		\vskip -0.2in
	\end{table*}	
	
\subsection{MULTIWAY TOPIC MODELING}
\label{sec:multiwaytopic}
\begin{figure}[!htbp]
		\begin{center}
			\centerline{
				\includegraphics[scale=0.25]{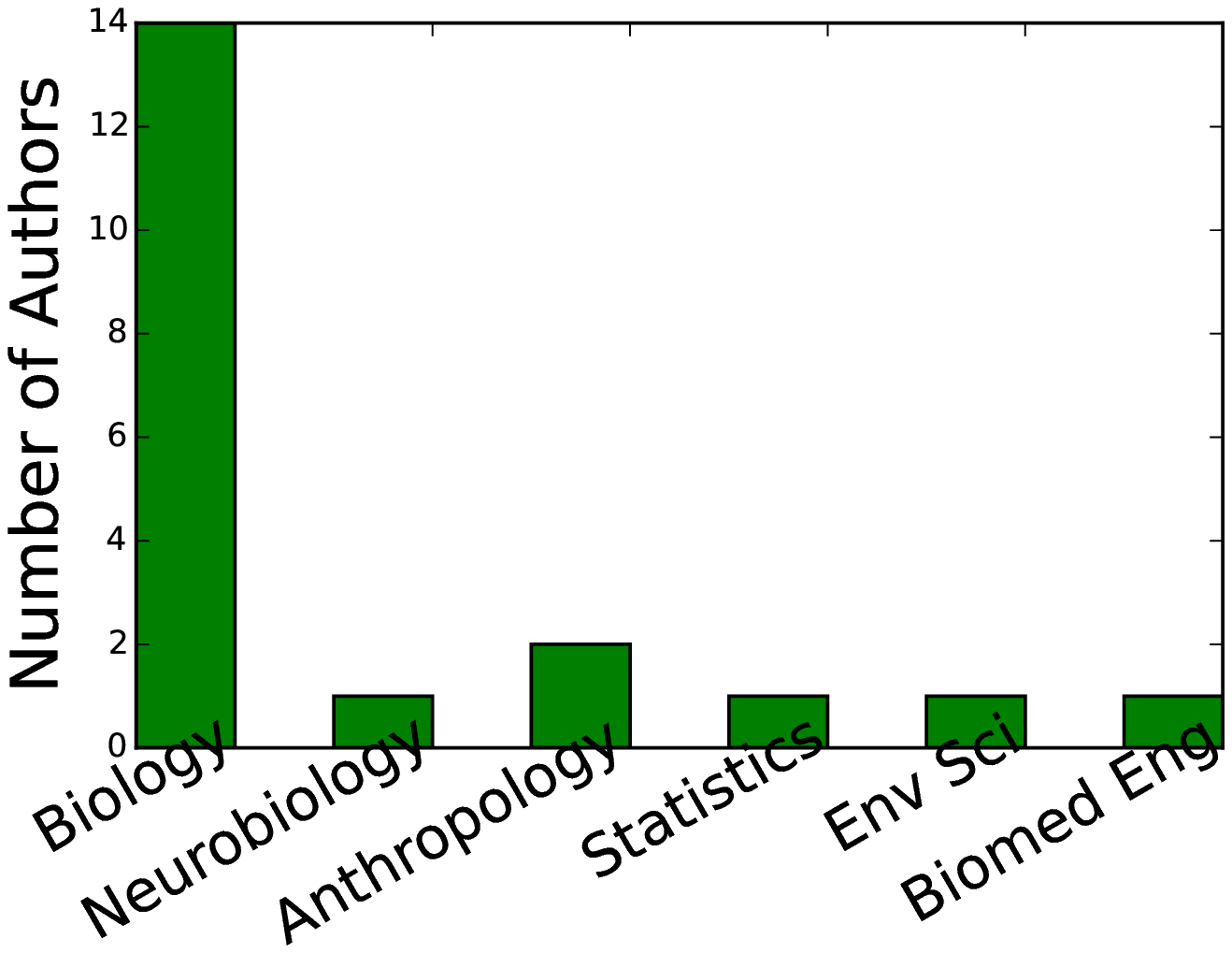}
				\includegraphics[scale=0.25]{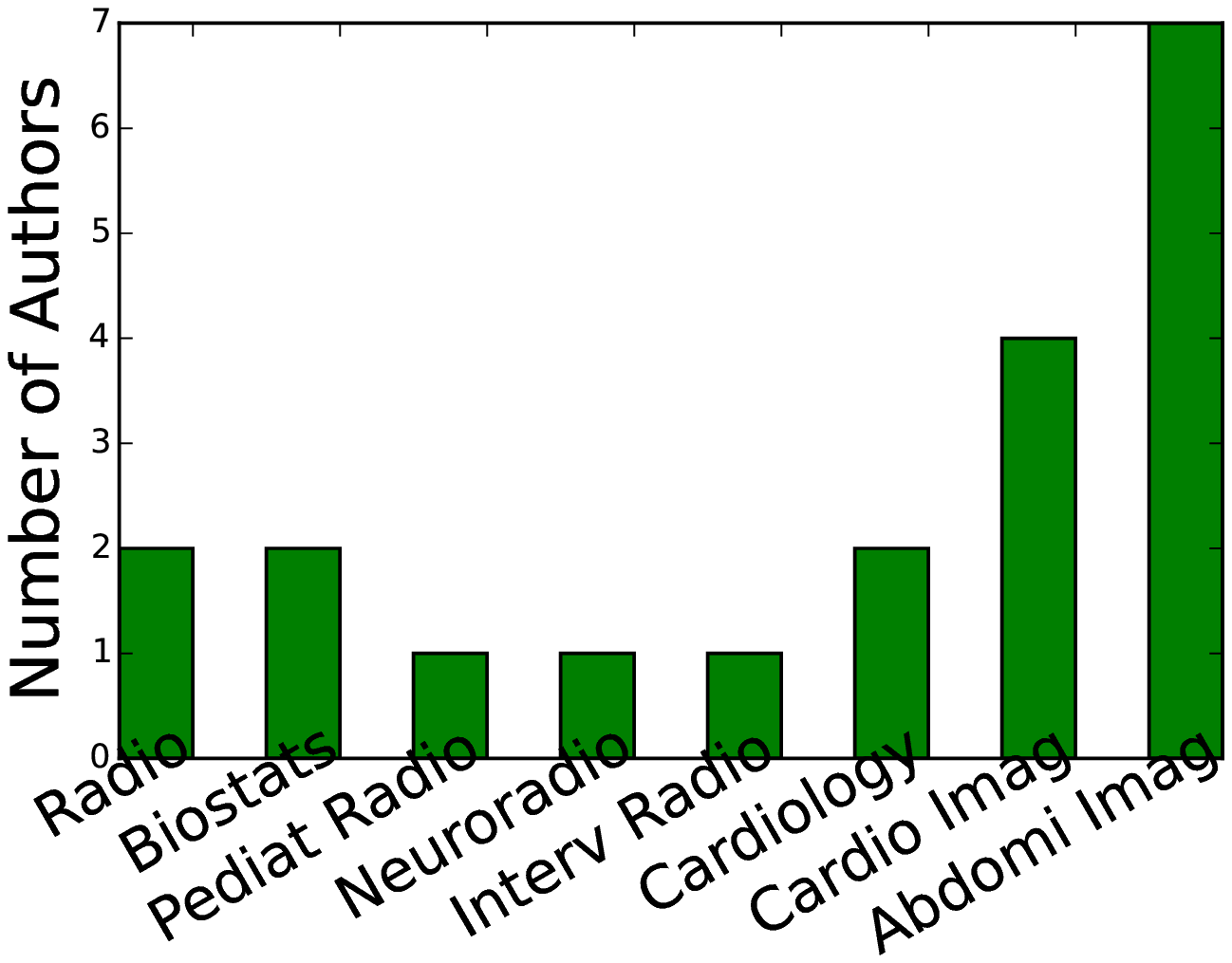}
			}
			\centerline{
				\includegraphics[scale=0.25]{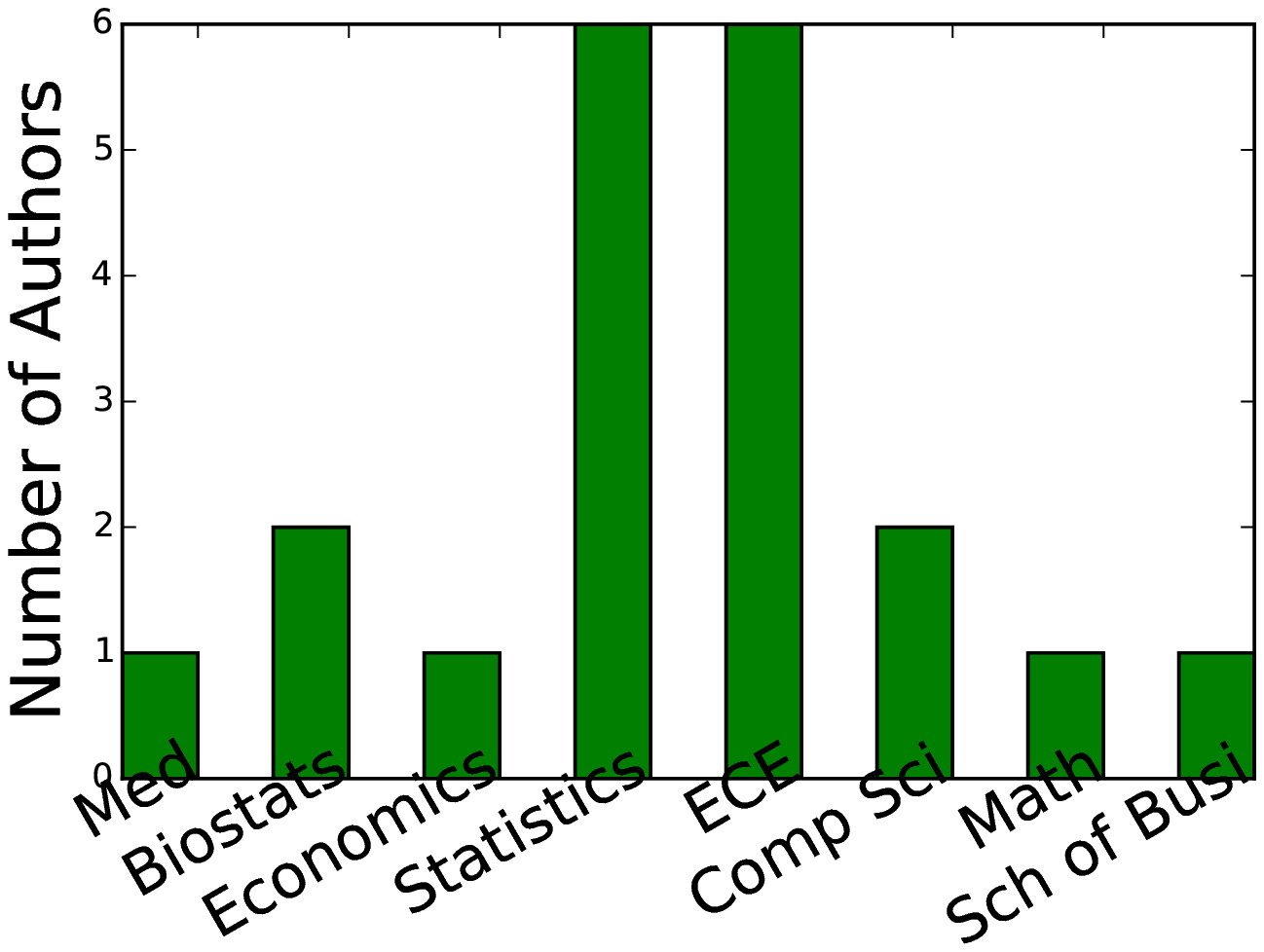}
				\includegraphics[scale=0.25]{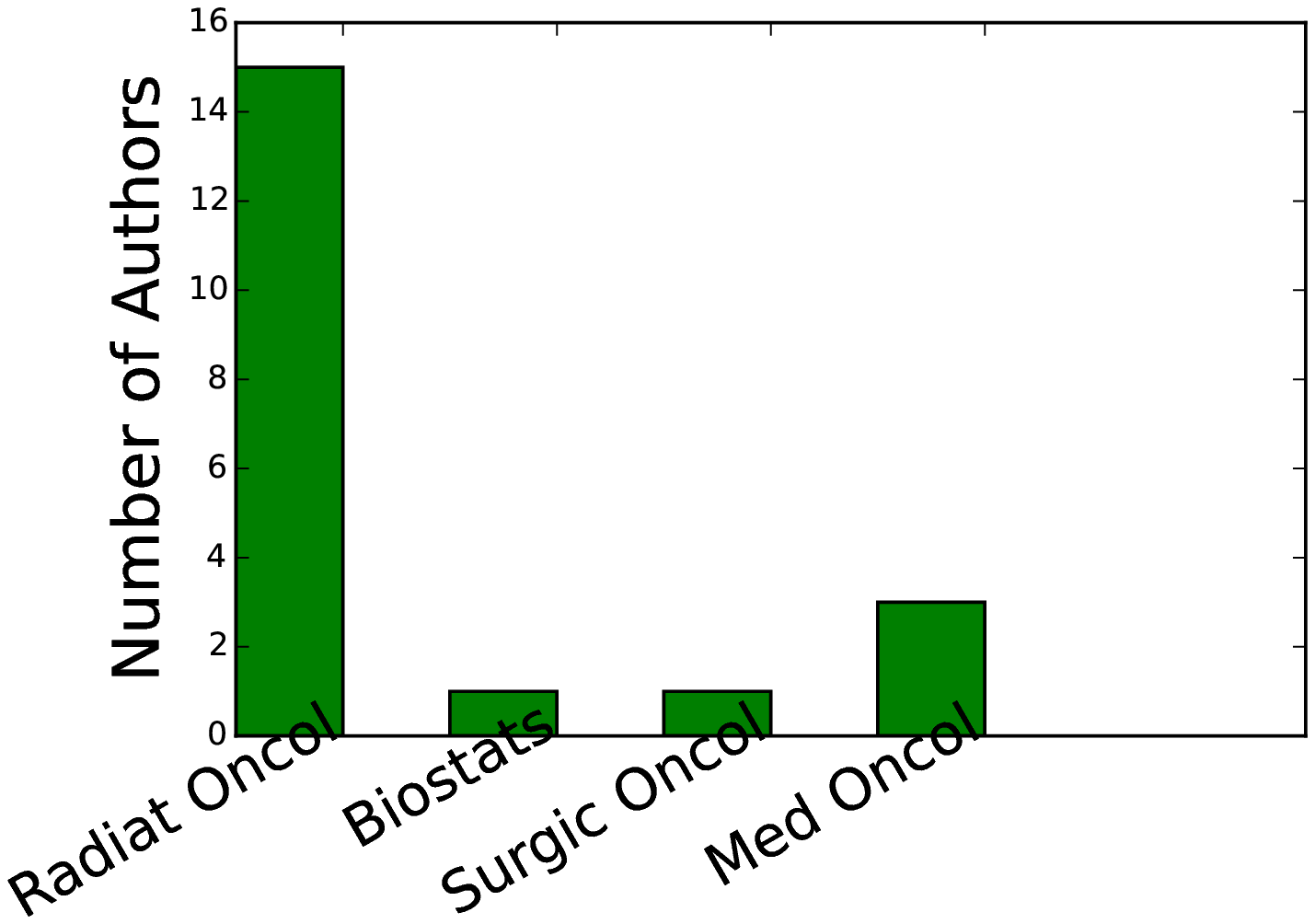}
			}
			\vskip -0.75em
			\caption{\small{Histogram of the department-affiliations for the top 20 authors in factors related to evolutionary biology (top left), medical imaging (top right), machine learning/signal processing(bottom left) and oncology (bottom right).}}
			\label{fig:hist}
		\end{center}
		\vskip -0.35in
	\end{figure}

We also apply our model for a \emph{multiway} topic modeling task on the Scholars data. The binary tensor represents $\textsc{authors} \times \textsc{words} \times \textsc{venues}$ relationships. We apply our model (with batch MCMC) and examine the latent factors of each of the three dimensions. Since each factor is drawn from a Dirichlet, it is non-negative and naturally corresponds to a ``topic''. In Table~\ref{tab:topic}, after examining the words factor matrix, we show the top-10 words for four of the factors (topics) inferred by our model; these factors seem to represent topics Evolutionary Biology, Medical Imaging, Machine Learning/Signal Processing, and Oncology. For the Machine Learning/Signal Processing topic, we also examine the corresponding topic in the venues factor matrix and show the top-10 venues in that topic (based on their factor scores in that factor). In Fig.~\ref{fig:hist}, we also show the histograms of authors' department affiliations for each of the four topics and the results make intuitive sense. The results in Table~\ref{tab:topic} and Fig.~\ref{fig:hist} demonstrate the usefulness of our model for scalable topic modeling of such multiway data.

\subsection{LEVERAGING THE MODE NETWORK}
\label{sec:tens-net}
\vspace{-0.5em}
Finally, to investigate the usefulness of leveraging the mode network, we experiment with using both the tensor \emph{and} the mode network on Scholars and Facebook data sets. For each data set, we report the AUC-ROC (area under the ROC curve) and AUC-PR (area under the precision-recall curve) on the tensor completion task, with and without network. For both data sets, we experiment with the more challenging cold-start setting. In particular, for the Facebook data, we hold out all the entries of the tensor slices after the first 50,000 wall-owners and predict those entries(using only the rest of the tensor, and using the rest of the tensor as well as the friendship network). We run the experiment with $R=20$ and minibatch size of 50,000 for the online MCMC. The results in Table~\ref{tab:cold} show that using the network leads to better tensor completion accuracies.

We also perform a similar experiment on the Scholars data where we hold out all the entries in tensor slices after the first 1000 authors and predict those entries (using only the rest of the tensor, and using the rest of the tensor as well as the co-authorship network). We run the experiment with $R=100$ and minibatch size of 50,000 for the online MCMC. The results shown in Table~\ref{tab:cold} again demonstrate the benefit of using the network. 

\vspace{-1.5em}
\begin{table}[!htbp]
\scriptsize
      \caption{\small{Cold-start setting}}
      \label{tab:cold}  
\center
      \begin{tabular}{|c|c|c|c|c|}
     	\hline				
				& \multicolumn{2}{c|}{Facebook} & \multicolumn{2}{c|}{Scholars}\\
				\hline
	  & \bf{AUC-ROC}  & \bf{AUC-PR} & \bf{AUC-ROC}  & \bf{AUC-PR}\\  
    	\hline    	
	\bf{Without network} & 0.8897 &  0.6076  & 0.8051 & 0.5763 \\ \hline
	\bf{With network} & \bf{0.9075} & \bf{0.7255} & \bf{0.8124} & \bf{0.6450}\\ \hline 
      \end{tabular}
			\vspace{-1em}
\end{table}

\begin{figure}[!htbp]
		\begin{center}
			\centerline{
				\includegraphics[scale=0.28]{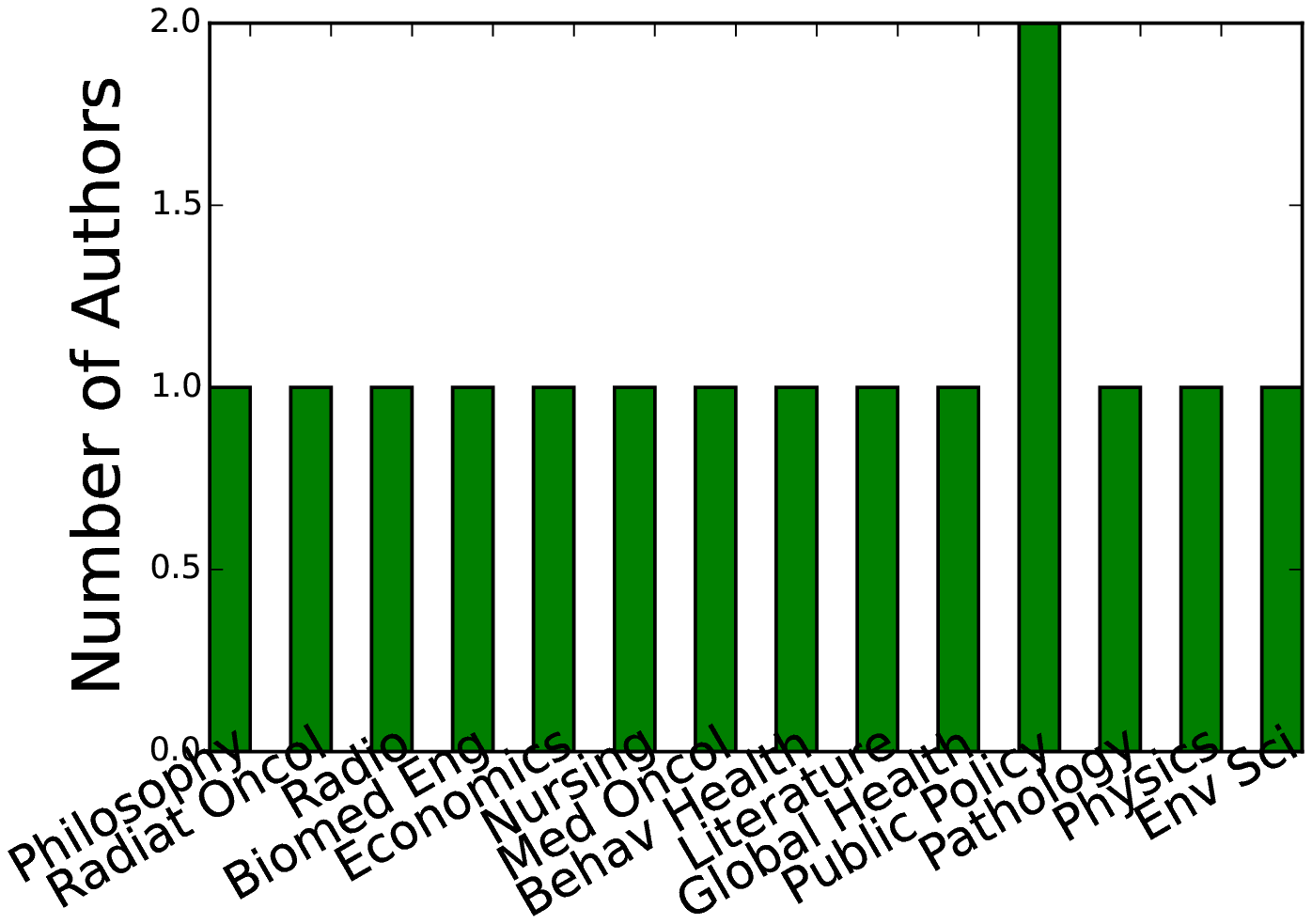}
				\includegraphics[scale=0.28]{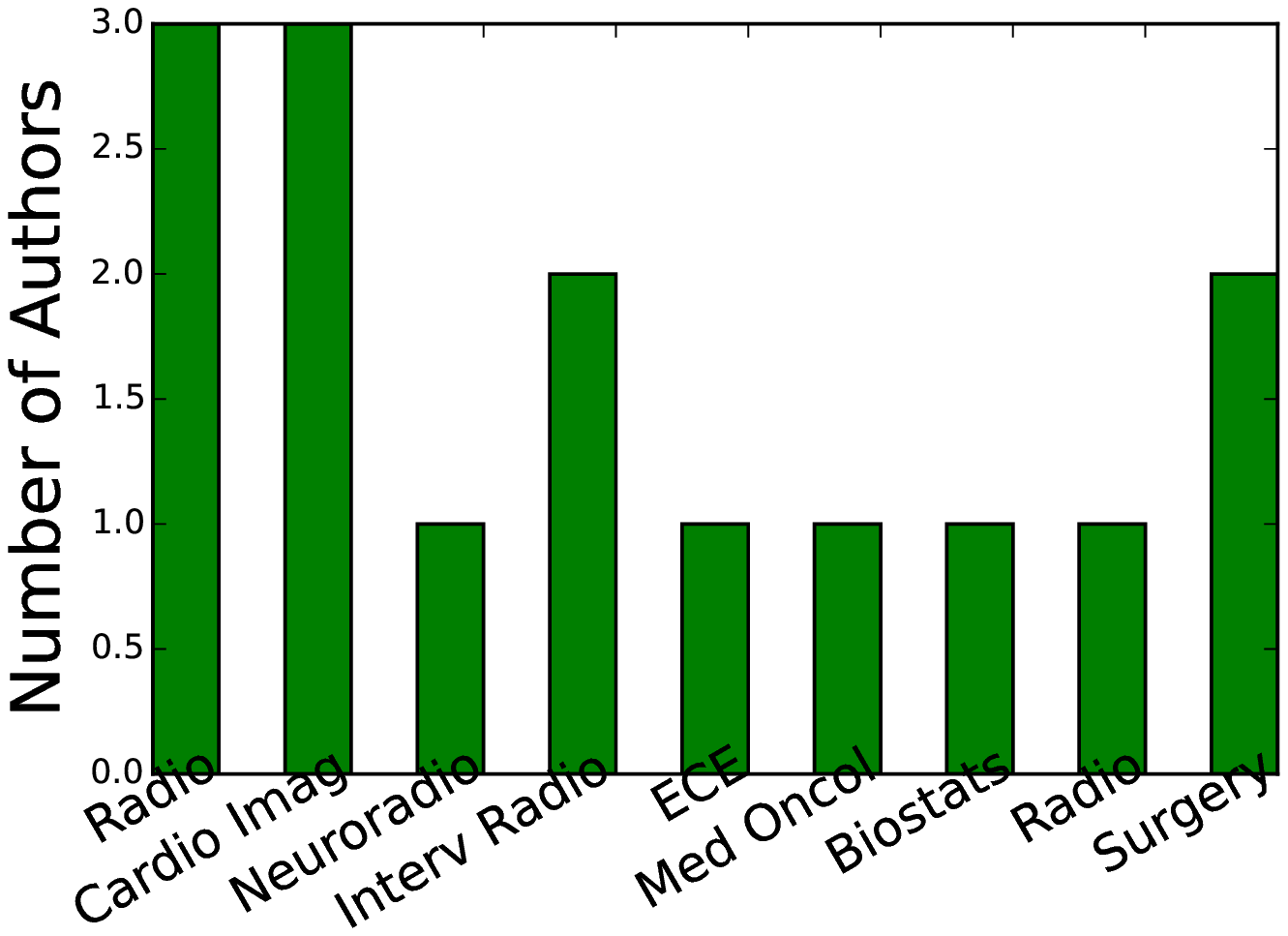}
			}
			\centerline{
				\includegraphics[scale=0.28]{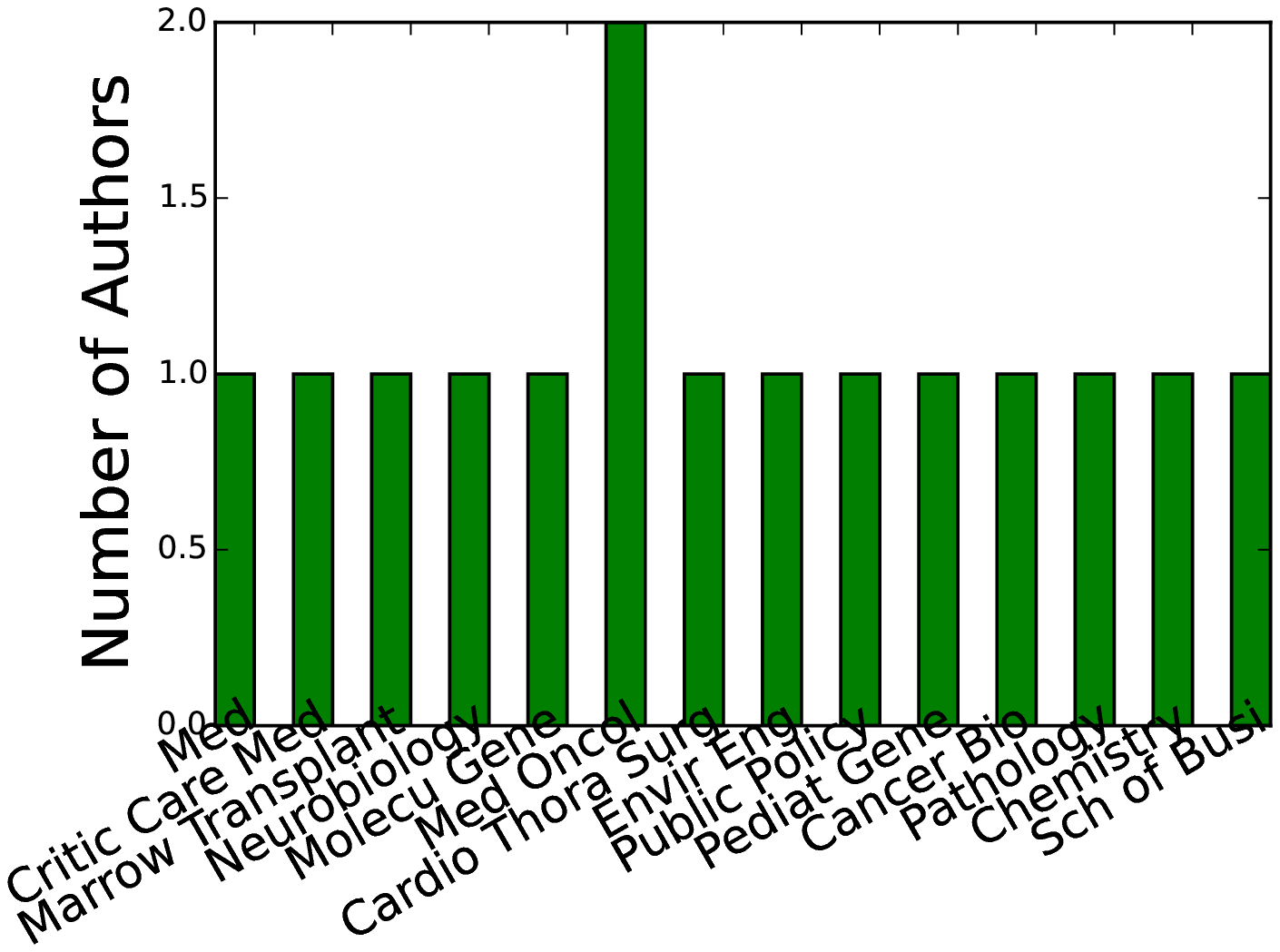}
				\includegraphics[scale=0.28]{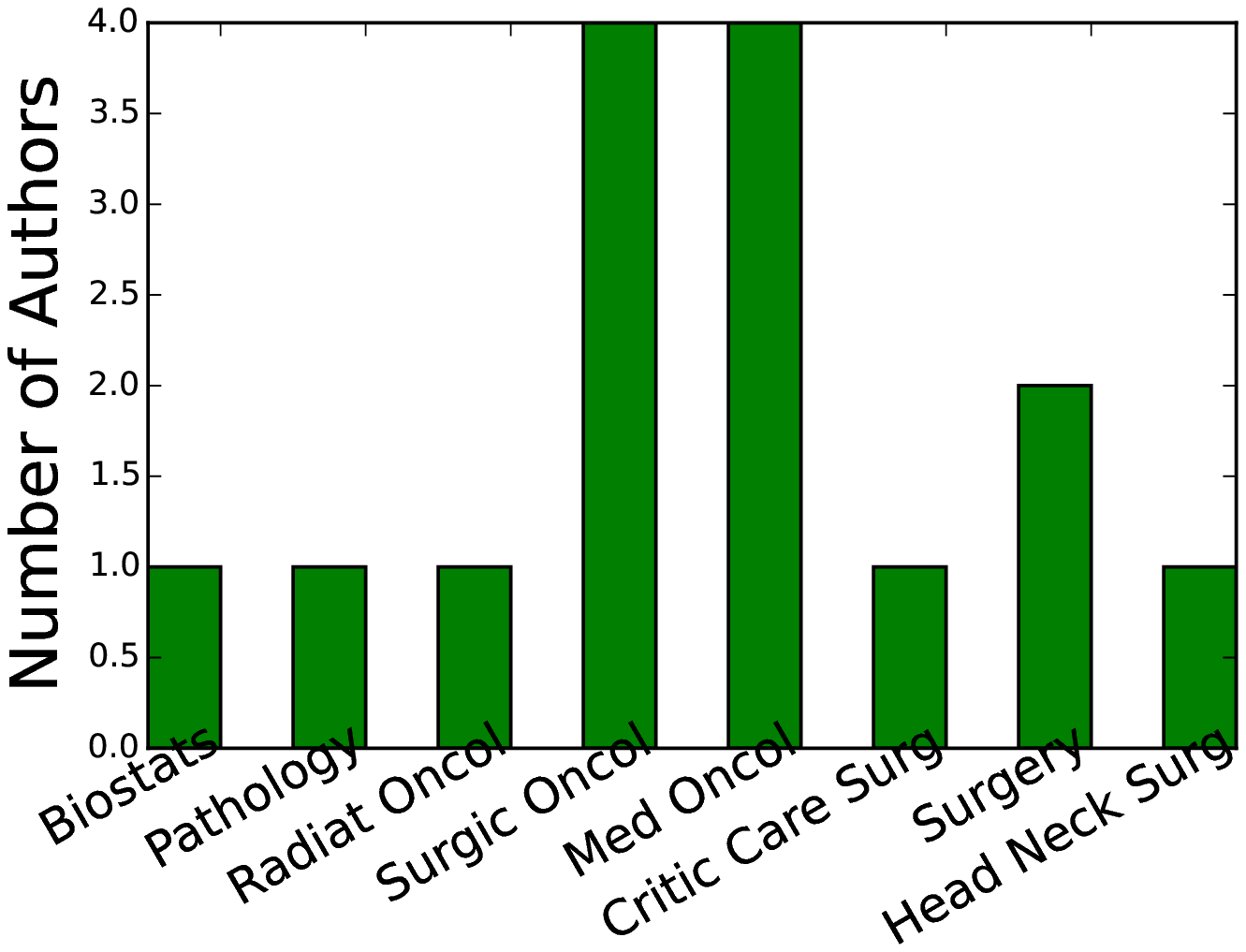}
			}
			\vskip -1em
			\caption{\small{Histogram of the department-affiliations of the top 15 \emph{held-out} authors associated with the factors of medical imaging (top) and oncology (bottom). The left column is obtained using no co-authorship information, and the right column is obtained using co-authorship information.}}
			\label{fig:sideinfohist}
		\end{center}
		\vspace{-2em}
	\end{figure}

In Fig.~\ref{fig:sideinfohist}, we show another result demonstrating the benefit of using the co-authorship network for the Scholars data. Note that in the cold-start setting, there is no information in the tensor for the \emph{held-out} authors. Therefore the topics associated with such authors are expected to be roughly \emph{uniformly random}. As shown in Fig.~\ref{fig:sideinfohist} (left column), the set of held-out authors assigned to the topics medical imaging and oncology seem very random and \emph{arbitrary} (we only show the aggregate department-affiliations). Using side-information (in form of the co-authorship network), however, the model sensibly assigns authors who are indeed related to these topics, as shown in right column of Fig.~\ref{fig:sideinfohist}.

%
%
%
\vspace{-1.5em}
\section{CONCLUSION}
\label{sec:concl}
\vspace{-1em}
We have presented a scalable Bayesian model for binary tensor factorization. In contrast to the models based on probit or logistic likelihood for binary tensor decomposition, the time-complexity of our model depends only in the number of ones in the tensor. This aspect of our model allows it to easily scale up to massive binary tensors. The simplicity of our model also leads to simple batch as well as online MCMC inference; the latter allows our model to scale up even when the number of ones could be massive. Our experimental results demonstrate that the model leads to speed-ups of an order of magnitude when compared to binary tensor factorization models based on the logistic likelihood, and also outperforms various other baselines. Our model also gives interpretable results which helps qualitative analysis of results. In addition, the ability to leverage mode networks (fully or partially observed) leads to improved tensor decomposition in cold-start problems. 
\vspace{-0.5em}
\section*{Acknowledgments}
\vspace{-1em}
The research reported here was supported in part by ARO, DARPA, DOE, NGA and ONR.

\bibliography{tensor}
\bibliographystyle{apalike}

\end{document}